\documentclass[runningheads]{llncs}
\usepackage{graphicx}
\usepackage{amsmath,amssymb} 
\usepackage{color}
\usepackage{amsmath}
\usepackage{tikz}
\usepackage{comment}
\usepackage{amsmath,amssymb} 

\usepackage{amsmath}
\usepackage{amssymb}
\usepackage{algorithm}
\usepackage{algpseudocode}
\usepackage{multirow}
\usepackage[accsupp]{axessibility}  

\begin{document}
\pagestyle{headings}
\mainmatter


\title{Deep Active Ensemble Sampling For Image Classification} 
\titlerunning{DAES}
\authorrunning{S. Mohamadi et al.}
\author{Salman Mohamadi, Gianfranco Doretto, Donald A. Adjeroh}
\institute{West Virginia University, Morgantown, WV, USA}

\maketitle

\begin{abstract}
Conventional active learning (AL) frameworks aim to reduce the cost of data annotation by actively requesting the labeling for the most informative data points. However, introducing AL to data hungry deep learning algorithms has been a challenge.
Some proposed approaches include uncertainty-based techniques, geometric methods, implicit combination of uncertainty-based and geometric approaches, and more  recently, frameworks based on  semi/self supervised techniques. 
In this paper, we address two specific problems in this area. The first is the need for efficient exploitation/exploration trade-off  in sample selection in AL. For this, we present an innovative integration of recent progress in both uncertainty-based and geometric frameworks to enable an efficient exploration/exploitation trade-off in sample selection strategy. 
  To this end, we build on a computationally efficient approximate of Thompson sampling with key changes as a posterior estimator for uncertainty representation. Our framework provides two  advantages: (1) accurate posterior estimation, and (2) tune-able trade-off between computational overhead and higher accuracy. The second problem is the need for improved training protocols in deep AL. For this, we use ideas from semi/self supervised learning to propose a general approach that is independent of the specific AL technique being used. 
   Taken these together, our framework shows a significant improvement over the state-of-the-art, with results that are comparable 
   to the performance of supervised-learning under the same setting. We show empirical results of our framework, and comparative performance with the state-of-the-art on four datasets, namely,  MNIST, CIFAR10, CIFAR100 and ImageNet to establish a new baseline in two different settings.
  
\end{abstract}
\section{Introduction}
Active learning (AL) has consistently played a central role in domains where labeling cost is of great concern. The core idea of AL frameworks revolves around learning from small amounts of annotated data and sequentially choosing the most informative data sample or batch of data samples to label. To this end, after initial training using available labeled data, an acquisition function is utilized to leverage the model's uncertainty in order to explore the pool of unlabeled data for most informative data points. In parallel with  advancements in AL, in the recent years, deep learning has gained tremendous attention with diverse applications such as realistic image generation, object detection and tracking, semantic segmentation, and iris recognition  \cite{khosravian2021generalizing,kashiani2019visual,mostofa2021deep} due to its  emergence as a high-performing approach, primarily conditioned on the availability of large amounts of training data. An interesting challenge is how to efficiently incorporate data-hungry deep learning tools into supposedly data-efficient AL frameworks.

Adjusting AL algorithms for deep neural networks has been 
very challenging, where extending the model complexity/capacity to that of CNNs ultimately ended up with either a poor performance, or some minor improvements at the cost of querying almost all samples. On the other hand, sequential training of such expressive models as well as extending the framework to high dimensional data injects even more complexity \cite{cohn1994improving,balcan2009agnostic,cesa2009robust}. This challenge was relatively under-explored, until a breakthrough work by Gal et al \cite{gal2017deep}, which essentially 
considered the problem of incorporating deep learning into AL for high dimensional data as highly connected with that of uncertainty representation \cite{mohamadi2020deep}. 
They thus approached the problem from the perspective of uncertainty representation in deep learning for AL, and 
 developed a Bayesian AL framework for image data. 
 Later work (such as \cite{beluch2018power}), however, argued that the approach exhibits poor scalability to big datasets due to its limited model capacity .
 
Another approach that also relied on  uncertainty representation, is  ensemble-based AL \cite{beluch2018power}. Here, an ensemble of classifiers is used, where the 
classifiers independently learn from the data in parallel. The major drawback is the poor diversity (lack of exploration) even with larger ensembles.
Our approach, while enjoying the power of ensembles, solves this problem by offering an inherent exploration/exploitation trade-off as classifiers maintain some dependency in the form of a shared prior.
Apart from uncertainty representation, another set of emerging methods  that primarily rely on geometrical data representation \cite{sener2018active} 
showed improved performance in deep AL. However, 
similar to \cite{sinha2019variational}, we empirically observed that these geometric approaches typically suffer from performance degradation as the class diversity (number of classes) increases. Another recent approach is the work reported in 
\cite{sinha2019variational} where they take advantage of adversarial training to provide improved performance over previous methods.
We empirically find that their work provided a balanced performance on datasets at different scales and diversity. As we will show later, our proposed model outperforms this approach in multiple settings with significant margins, with results approaching that of supervised learning models in some cases. 
 
 In the first part of the paper, primarily motivated to efficiently 
 integrate the advantages of uncertainty and geometrical 
 representations,  we propose an approach built upon approximate Thompson sampling. On one hand, this provides an improved representation of uncertainty over unlabeled data, and on the other hand, supports an  inherent tune-able exploration/exploitation trade-off 
 for diverse sampling \cite{bouneffouf2014contextual,ganti2013building}. 
 Unlike conventional ensemble-based methods whose performance tend to saturate quickly, under our tuneable model, adding a few more classifiers tends to improve
 the uncertainty and geometric representation.
To mitigate the general sample diversity problem of ensemble models (see \cite{melville2004diverse,beluch2018power} ), we use an inclusive sample selection strategy. Our framework showed a noticeable improvement over the state-of-the-art, with performance approaching those of supervised learning methods. Further, we explore the scope and scale of model efficiency improvements brought about by our proposed  techniques.  

Briefly, due to the exploration/exploitation trade-off, Thompson sampling is expected to improve both 
predictive uncertainty and sample diversity  by computing, sampling, and updating a posterior distribution.
A serious consideration, however, is that, for more expressive models such as deep  convolutional neural networks (CNNs) designed for high dimensional data, Thompson sampling makes the process 
computationally difficult. This is primarily because computation of the posterior distribution over CNNs is complex by nature. 
Inferences based on Laplace approximations or Markov chain Monte Carlo approaches would be two possible 
alternatives. 
However, both approaches are still very  expensive in terms of computational cost \cite{chapelle2011empirical,brooks2011handbook,lu2017ensemble}. 
Lu et al \cite{lu2017ensemble} argue that due to the compatibility of Thompson sampling with sequential decision and updating, 
an approximate version of Thompson sampling  
could be a promising solution.
Accordingly, we build an ensemble model relying on an efficient approximate of Thompson sampling, which improves the state-of-the-art. 
Interestingly, this model possesses  both the advantage of uncertainty based deep AL 
approaches (exploiting most uncertain samples), and 
of geometric solutions (exploring for more diverse though not necessarily highly uncertain samples).

In the second part of the paper, we investigate a new line of efforts/arguments revolving around the idea of boosting AL frameworks using self/semi supervised learning techniques. We substantiate and unify these arguments 
and also design and perform extensive experiments on multiple baselines to assess 
this approach as a new general training protocol for AL frameworks. This enables our approach to be compared against recent boosted AL frameworks. 

Briefly, our key contributions in this paper are as follows:
\begin{itemize}
    \item A new framework for deep AL which enables an exploration/exploitation trade-off for sample selection and hence offers the advantages of both uncertainty-based and geometry-based methods.
    \item 
     A new general training protocol for visual AL approaches, developed by substantiating and unifying recent arguments on boosting AL using self/semi supervised learning, and experimentally evaluating this approach on multiple recent baselines.  
    We compare our framework against two sets of baselines to show its performance.
\end{itemize}
\section{Background and preliminaries}
\textbf{Background:} Early efforts on AL with image data 
considered mainly kernel-based approaches \cite{zhu2003combining,li2013adaptive,joshi2009multi}.
Later, 
AL methods with image data using CNN included uncertainty-based approaches \cite{gal2017deep,gal2016dropout,beluch2018power,lakshminarayanan2016simple,osband2016deep}, geometry-based approaches \cite{sener2018active}, or their 
combination  \cite{sinha2019variational}, e.g, based on adversarial training. Generally speaking, uncertainty-based approaches focus on finding most uncertain samples to label, with the potential downside of less diversity in sample selection, while geometric approaches tend to weigh on diversity of samples, resulting in performance degradation in cases of very diverse datasets (with large number of classes).
Most recently, in a relatively different setting, Gao et al. \cite{gao2020consistency} leveraged semi-supervised learning while Bengar et al. \cite{bengar2021reducing} applied self-supervised learning (SSL) techniques to deliver a significant performance improvement.  We will compare our proposed approach against these related work, on the same problem settings. 
Some other recent work in this general area of modern AL with high dimensional data can be found in \cite{yoo2019learning,agarwal2020contextual,cortes2020adaptive,zhang2021datasetgan,zhang2020state,ebrahimi2020minimax}. Though these are relevant, they are not as closely related to our approach.\\
\textbf{SSL:}  As the second contribution of this work relates to SSL we briefly review the literature. Briefly, SSL is one of the closest modern problem domains to AL with zero labeling effort policy. Here, the goal is to leverage all unlabeled data to train a network for a pretext task so as to  prepare the network  for a downstream task, usually with small amounts of data \cite{jing2021self}. Until recently, a major set of SSL baselines were contrastive baselines relying on contrasting augmented views of a sample with each other (positive contrastive pairs) and with views of other samples (negative contrastive pairs) \cite{hadsell2006dimensionality,wu2018unsupervised}. Newer  baselines such as \cite{grill2020bootstrap,chen2021exploring}, a.k.a non-contrastive approaches, rely on contrasting positive pairs, needless of contrasting negative pairs. 
Recently, Ermolov et al. \cite{ermolov2021whitening} reported a non-contrastive method based on whitening the embedding space, which was effective, yet conceptually simple. We adopt this approach in this work.\\ 
\textbf{Preliminary:} We describe these two major paradigms below.\\
\textbf{1. Uncertainty-based techniques:}
Two categories of well-known deep learning techniques for uncertainty representation and estimation include ensemble-based techniques (non-Bayesian) \cite{lakshminarayanan2016simple,osband2016deep} and Monte-Carlo (MC) dropout (Bayesian) \cite{gal2016dropout,gal2017deep}. 
In ensemble-based methods, an ensemble of ${N}$ identically structured neural networks are trained using identical training data ${D_{tr}}$, where the different random values are applied for weight initialization $w_i$. 
For a given class $c$ out of multiple classes and input $X$, we then have:
\begin{equation}\begin{split}
p(y=c|x,D_{tr})=\frac{1}{N}\sum_{i=1}^{i=N}p(y=c|x,w_i)
\end{split}
\label{eq:010}
\end{equation}

However, MC-dropout 
trains a network with dropout, and during test, implements $T$ forward passes, each individually with a new dropout mask, resulting in $T$ sets of weights $w_t$. Given input $x$, the average of all $T$ softmax vectors represents the output for a desired class $c$.
\begin{equation}\begin{split}
p(y=c|x,D_{tr})=\frac{1}{T}\sum_{t=1}^{t=T}p(y=c|x,w_t)
\end{split}
\label{eq:020}
\end{equation}
Here we briefly describe some popular effective uncertainty-based acquisition functions \cite{gal2017deep,beluch2018power} or their approximation for ensemble-based approaches, MC dropout, and 
our proposed framework, all  based on uncertainty sampling.


\textbf{A.} Selecting samples with highest predictive entropy \cite{shannon1948mathematical}.
\begin{equation}\begin{split}
H[y|x, D_{tr}]:=-\sum_{c}(\frac{1}{N}\sum_{n}p(y=c|x,w_n))
.\log(\frac{1}{N}\sum_{n}p(y=c|x,w_n))
\end{split}
\label{eq:0100}
\end{equation}

\textbf{B.} Selecting samples with highest mutual information between their predicted labels and the weights, BALD \cite{houlsby2011bayesian,gal2017deep}, which was initially applied in \cite{gal2017deep} with $T$ forward passes in MC-dropout. It can be analogously rewritten for an ensemble with $N$ members by replacing $T$ with $N$.
\begin{equation}\begin{split}
I[y;w|x, D_{tr}]:=H[y|x,D_{tr}]
-\frac{1}{T}\sum_{t}\sum_{c}-p(y=c|x,w_t).\log p(y=c|x.w_t)
\end{split}
\label{eq:01000}
\end{equation}

\textbf{C.} Highest Variation Ratio \cite{freeman1965elementary} as a measurement of non-modal  predicted class labels, where $f_m$ is the number of modal class predictions \cite{beluch2018power}.
\begin{equation}\begin{split}
VR:=1-f_m /N 
\end{split}
\label{eq:01000}
\end{equation}
We used this acquisition function in our proposed  DAES framework.\\
\textbf{2. Geometry-based techniques:}
Geometric or representation-based methods primarily rely on density-based acquisition functions. Typical examples include REPR \cite{yang2017suggestive}, and Core-Set \cite{sener2017geometric}. With a total of $n$ samples, at each iteration Core-Set selects a fixed number of samples, that minimize the upper bound on the distance between point $x_i$ in $n$ samples, and $x_j$, its closest neighbour in selected subset $o$. The acquisition function of Core-Set is given as follows:
$ s= argmax_{i\in [n]\o} \min_{j\in o} dist(x_i,x_j) $. See \cite{yang2017suggestive} for that of REPR.\\
\textbf{3. Other techniques:}\\
Other methods include implicit combination of uncertainty and geometry approaches, such as in \cite{sinha2019variational}, which designs a minimax game in the context of adversarial training. There are also methods that have used  the power of pre-trained models such as \cite{gao2020consistency}, and to a less extent \cite{guo2021semi}.
\section{Deep active ensemble sampling}
Our work is primarily inspired by the reports in \cite{gal2017deep,osband2016deep,lu2017ensemble} towards finding an uncertainty-diversity trade-off. In particular, we propose a tuneable trade-off between uncertainty-wise exploitation of samples vs exploration of less uncertain, but more diverse samples.
\subsection{Thompson sampling for AL}
\textbf{Contextual bandit:} Thompson sampling was primarily developed as a heuristic to address the Multi-armed bandit (MAB) problem, aiming for a trade-off between exploration and exploitation in sequential decision making. 
The core idea of Thompson sampling has a Bayesian essence (See Algorithm 1). 
Unlike greedy algorithms that mostly lean toward exploitation, Thompson sampling  draws random samples from a posterior distribution to fine-tune between exploration and exploitation.  
See \cite{ganti2013building,bouneffouf2014contextual} for related work on low dimensional data.
%
New attempts towards using Thompson sampling for efficient estimation of posterior distribution for more complex models such as CNNs revealed an immediate need to find a computationally tractable approximation \cite{lu2017ensemble}.\\
\textbf{Deep AL:} Assuming a pool-based AL setting, we initially have a set of unannotated data ${U_0=\{x_1, x_2,... x_n\}}$ and a small set of annotated data ${A_0}$, where at each iteration, an algorithm, known as \textit{acquisition function}, looks into the whole set of unlabeled data to select a number of samples and pass them to an Oracle for labeling. In deep learning backed AL with high dimensional data such as images, the goal is to adjust the model to enable learning from a relatively small initial training set, and accordingly
select a subset of most informative unlabeled data samples (in terms of uncertainty and diversity) to be labeled.

\subsection{Ensemble sampling}
From a geometry perspective, one ideal estimation of the desired posterior space in AL framework could be represented by a direct sum over the space. Along this line, 
some methods such as \cite{cortes2020adaptive} propose splitting the input space to  improve uncertainty sampling associated with the posterior distribution. Hinton et al \cite{hinton1999unsupervised} noted  the fact that data points are generated by natural sources that actually inject limited complexity, rather than random sources with unlimited complexity. Therefore, unlike a random source that practically enables sampling from an infinite space, the natural source can be represented with a direct sum over the posterior space ${ \mathbf{S}}$ with any \textbf{\textit{finite}} number of summands ${Q}$: ${\mathbf{S}=\mathbf{S_1}\oplus\mathbf{S_2}\oplus ... \oplus\mathbf{S_Q}}$, where ${\mathbf{S_i}}$ represents the $i$-th subspace. 
Later we will see that compared with regular ensembles, ensemble sampling is closer to this direct sum as it  allows a better exploration of whole representation space. 

In the case of AL on a neural network with weights ${\theta}$, let's say the network represents the mapping ${g_{\theta}:\mathcal{R}^W\mapsto \mathcal{R}^K}$ (W is the dimensionality of input) and the goal is to sequentially choose a fixed number of samples ${d_t}$ from a pool ${\mathcal{D}}$ of ${K}$ samples as input at each time ${t=0, 1, ... T }$,  where ${\mathcal{D} \subseteq \mathcal{R}^W}$, such that it leads to desirable output. Accordingly, with each set of samples ${d_t}$ selected from ${\mathcal{D}}$ at time ${t=0, 1, ... ,T}$, an output ${g_\theta(d_t)}$ and random variable  
${w_{t}\sim N(0,\sigma_w ^2 I)}$ form the observation ${y_t=g_\theta(d_t)+w_t}$ which allows to update a reward ${r_t=r(w_t)}$ sequentially. Supposing that we have a prior on ${\theta}$, ${\theta\sim N(\mu_0,\Sigma_0)}$, the model will become much more prone to uncertainty. Therefore, at each time ${t}$, the neural network will be fitted by ${d_t,y_t}$, and the samples are selected with the goal of converging to a trade off between immediate desirable outputs (minimizing the loss) and reducing uncertainty in ${\theta}$.

With the problem presented in as above, an algorithm is required to incorporate Thompson sampling in this new context. In the case of linear bandit problem, since the conventional Thompson sampling yields an efficient solution, no approximation to Thompson sampling is needed. However, in case of neural networks, the conventional form of Thomson sampling could be computationally expensive. This calls for a more efficient implementation in terms of approximate Thompson sampling. Accordingly, Lu et. al \cite{lu2017ensemble} introduce an ensemble of $N$ networks with a shared prior on their weights, as an approximate Thompson sampling. This allows efficient posterior estimation on complex models such as neural network. 
\subsection{Algorithms for CNNs}
Here we represent ensemble sampling as an efficient approximation of Thompson sampling for neural networks. In fact, unlike in simpler cases such as linear bandit, exact Bayesian inference can not easily be performed effectively for neural networks, which necessitates an efficient approximation. First, we present the algorithm for Thompson sampling (Algorithm 1 (taken from \cite{russo2018vko_TS}) ). Then, we discuss ensemble sampling as its efficient approximation, and present the algorithm for Deep Active Ensemble Sampling (Algorithm 2).

 More precisely on Thompson sampling, let's assume ${\mathcal{X}}$ is a finite set of data points ${x_1, ..x_n}$, where selecting a data point ${x_t}$ (or a number of data points) at time $t$ yields a randomly generated output ${y_t}$ based on a conditional probability distribution $q(.|x_t)$. Accordingly, a known function $r_t=r(y_t)$ is defined to capture the reward for the selected data point. This reward can be interpreted as a negative loss. At the beginning, the decision maker gets initialized with a prior $p$ on $\theta$, and as it starts to explore, updates its uncertainty representation. While greedy algorithms generally use expected value of ${\theta}$ with respect to ${p}$ to produce model parameters ${\hat{\theta}}$, Thompson sampling relies on random sampling from ${p}$. Next, the algorithm will choose data points maximizing the expected reward presented as follows:
\begin{equation}\begin{split}
\mathbb{E}_{q_{\hat{\theta}}}[r(y_t)|x_t=x]=\sum_{o}q_{\hat{\theta}}(o|x)r(o)
\end{split}
\label{eq:1}
\end{equation} 
Subsequently ${p}$ is updated by conditioning on ${\hat{y_t}}$, and for ${\theta}$ coming from a finite set, relying on Bayes rule we will have:
\begin{equation}\begin{split}
\mathbb{P}_{p,q}(\theta=u|x_t,y_t)=\frac{p(u)q_u(y_t|x_t)}{\sum_v p(v)q_v(y_t|x_t)}
\end{split}
\label{eq:2}
\end{equation}
Algorithm 1 (taken from \cite{russo2018vko_TS}) captures the above steps. As noted, this will be very time consuming, especially for neural networks. 

\begin{algorithm}
\caption{ Thompson(${\mathcal{X},p,q,r}$)}\label{alg:cap}
\begin{algorithmic}[1]
\For{$t$= 1, 2, ... , $T$ }
\State Sample $\hat{\theta}\sim p$
\State ${x_t \leftarrow \arg\max_{x\in \mathcal{X}} \mathbb{E}_{q_{\hat{\theta}}}[r(y_t)|x_t=x ]}$
\State {Input chosen $x_t$ and observe $y_t$}
\State ${p \leftarrow \mathbb{P}_{p,q}(\theta \in .| x_t, y_t)}$
\EndFor
\end{algorithmic}
\end{algorithm}

As an efficient approximate Thompson sampling for neural networks, we use ensemble sampling, where we employ an ensemble of $M$ networks and set priors on the weights, as presented in Algorithm 2. All networks will be trained on identical data samples while the initial shared priors on the weights makes a connection between them. 
Algorithm 2 is inspired by \cite{lu2017ensemble}, with key adjustments to make the approximate Thompson sampling adaptable to the AL framework. 
These changes include  (1) the optimization process of ensembles; (2)  selecting a set of samples rather than one sample; (3) we replace the original concept of maximizing reward in the algorithm with minimizing the loss, namely, 
$\Bar{L}({\theta})$ 
in our deep active learning framework. Accordingly, it is 
important to mention that the optimization of the method need not to be combinatorial as in the case with combinatorial contextual bandits. Moreover, sample selection is sequential in which, each iteration of sample selection provides a batch of samples ranked by the acquisition function. Unlike classical ensemble-based approaches, the proposed deep active ensemble sampling (DAES) not only puts a joint prior on the weights of the networks (all sampled from one prior distribution rather than individual priors), but also jointly optimizes the members of an ensemble.

 \begin{algorithm}
\caption{Deep Active Ensemble Sampling (M)}\label{alg:cap}
\begin{algorithmic}[1]
\State {Ensemble $En_M\left(g, \mathcal{N}(\mu,\,\sigma^{2})\right)$: $g({\theta_{1}}), ...,g({\theta_{M}})$; Labeled Set: $S^t_l$; Unlabeled Set: $S^t_u$}
\For{\textit{t}= 1, 2, ... , \textit{T} }
\State Train over $S_l^t$: {$En_M$: $g({\theta_{1,t}}), ...,g({\theta_{M,t}})$}

\State Optimize: ${{argmin}_{\theta_{i,t}}
 (L^t)= {argmin}_{\theta_{i,t}} \left(L(\theta_{1,t})+ ... + L(\theta_{M,t})\right)}$

\State Batch $b^t$ selection by fixed $En_M$: {$En_M(S^t_u)$, $VR=(1-\frac{f_m}{M})$} via Eqn (\ref{eq:01000})

\State Update Training Set: $S_l^{t+1}= S_l^{t}+b^t$
\EndFor
\end{algorithmic}
\end{algorithm}

\subsection{DAES with self-trained knowledge distillation}
Consistent with primary focus of AL on less annotation effort and with the goal of establishing a new standard AL training protocol, we empirically evaluate a simple training technique which inherently empowers any active learner, regardless of the underlying approach. While this is inspired by the recent trend in \cite{gao2020consistency,bengar2021reducing}, we also argue that using pre-training, here SSL pre-training, enables any AL framework to better model uncertainty over the data, or to capture the geometry of the data, due to the prior knowledge attained by SSL.
To ensure  fairness of our comparisons, we apply the new training protocol to both  the previous baseline AL models, and to our proposed DAES framework.
The proposed training protocol  could help to eventually unify this line of work with some form of knowledge distillation \cite{xu2020knowledge,bhat2021distill}.

\begin{figure}
    \centering
    \includegraphics[scale=.27]{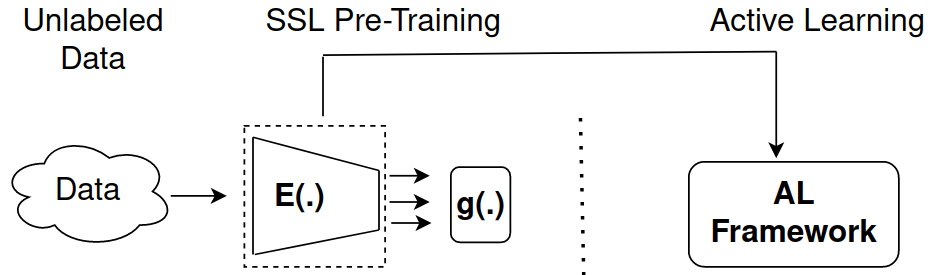}
    \caption{
    SSL pre-training for deep active learning. Here, $E(.)$ is the encoder and $g(.)$ is the projection head.
    After pre-training, the weights of $E(.)$ will be fixed and will then be used in our AL setting, training a classifier head on top of that.}
    \label{fig:my_SSL}
\end{figure}

\textbf{Training protocol: }
The protocol is a two step process: SSL pre-training, and then active learning using  the pre-training output. (See Fig. \ref{fig:my_SSL}).
Due to huge success of SSL in learning representation from unlabeled data, we adopt a most recent SSL model suitable for our setting.
Thus, our proposal for training AL models is to  consider a training protocol, first a pre-training is performed on the deep network (encoder in Fig. \ref{fig:my_SSL}) as a building block for the active learning models.
In this work, we tested this idea by adopting the conceptually simple, yet  effective SSL model in \cite{ermolov2021whitening} to initially train  ResNet18 as the building block for the AL methods, namely, Random baseline, VAAL, Core-Set and DAES.

We explore a new setting in which a given baseline is equipped with a conceptually simple self-training as discussed above.
As shown in Fig. \ref{fig:my_SSL}, we adopt the SSL framework from \cite{ermolov2021whitening}, to leverage knowledge distilled from unlabeled data for empowering the active learner. The idea is to use whitening in SSL in order to train the encoder (ResNet18) and then freeze all layers except for head-layers which are replaced with fully connected layers to be trained.
\section{Experiments and results}
We conduct two sets of experiments on images classification task to evaluate our proposed DAES framework as well as compare it against state-of-the-art models. Specifically, we mainly perform the experiments on MNIST \cite{lecun1998mnist}, CIFAR10 and CIFAR100 \cite{krizhevsky2009learning}, and ImageNet \cite{deng2009imagenet}. To ensure the fairness of compassion scenarios, we compare the framework against \textbf{two sets} of baselines, namely, trained from scratch, and self-trained enabled by self/semi supervised learning (SSL). \\
\textbf{Evaluation:} On CIFAR10/100 and ImageNet, starting with an initial budget of $10\%$ labeled samples, we measure the performance on sequential training using  $T$ training iterations, where in each iteration of training we add ${5\%}$ labeled data from unlabeled pool to the training set (labeled data ratio of $0.1$, $0.15$, $0.20$, ... up to $0.35$ or $0.50$). We assume each training iteration is from scratch unless otherwise stated. On MNIST the initial training set is 200 samples and the evaluation is performed on acquisition budget of 100 samples.
The results of all our experiments on all datasets including ImageNet are averaged over three trials.\\
\textbf{Baselines:} We compare the performance of DAES against two sets of baselines. First set of approaches, specifically trained from scratch, includes Random sampling from unlabeled pool (Random), Monte-Carlo dropout (M-C Dropout) \cite{gal2016dropout}, deep Bayesian active learning (DBAL) \cite{gal2017deep}, Core-Set \cite{sener2018active}, Ensemble with Variation Ratio (Ens-VarR) \cite{beluch2018power}, and VAAL \cite{sinha2019variational}. We also design and implement another set of extensive experiments on our framework as well as some of previous baselines empowered by self-training including Random, Core-Set, and VAAL to contrast against a very recent baseline taking advantage of SSL, CSSAL \cite{gao2020consistency}, and also later compare with a semi-supervised baseline, REVIVA \cite{guo2021semi}.

\subsection{Experimental settings}
We implemented our network architectures in Pytorch. 
Besides our experiments, experiments of all other competitive baselines including Random baseline, on CIFAR10, CIFAR100 and ImageNet are performed with ResNet18, with similar setting of VAAL except they used VGG16 \cite{simonyan2014very}. However for MNIST, we used a three-layer (two convolutional and one fully connected) network described in \cite{gal2017deep}. Specifically an ensemble includes $N=5$ identical classifiers unless otherwise specified. 
We used Xavier initialization when applicable, and we utilized Adam optimizer \cite{kingma2014adam} for all experiments. All experiments start with an initial balanced budget of $10\%$ of unlabeled training pool (6000 for MNIST, 5000 for CIFAR10/100, and 128120 for ImageNet), which is then iteratively updated by adding $5\%$ of whole training pool. Both initial training and other sequential iterations of training continue for 100 epochs. After every update, the network is trained from scratch unless otherwise specified (i.e., incremental training). Further, unlike classical ensemble-based methods, the optimization process of all classifiers in DAES is performed jointly as one loss function. Practical considerations in case of DAES with very deep networks are discussed in ablation studies.\\
\begin{figure*}[t]
	\centering{
		\includegraphics[scale=.27]{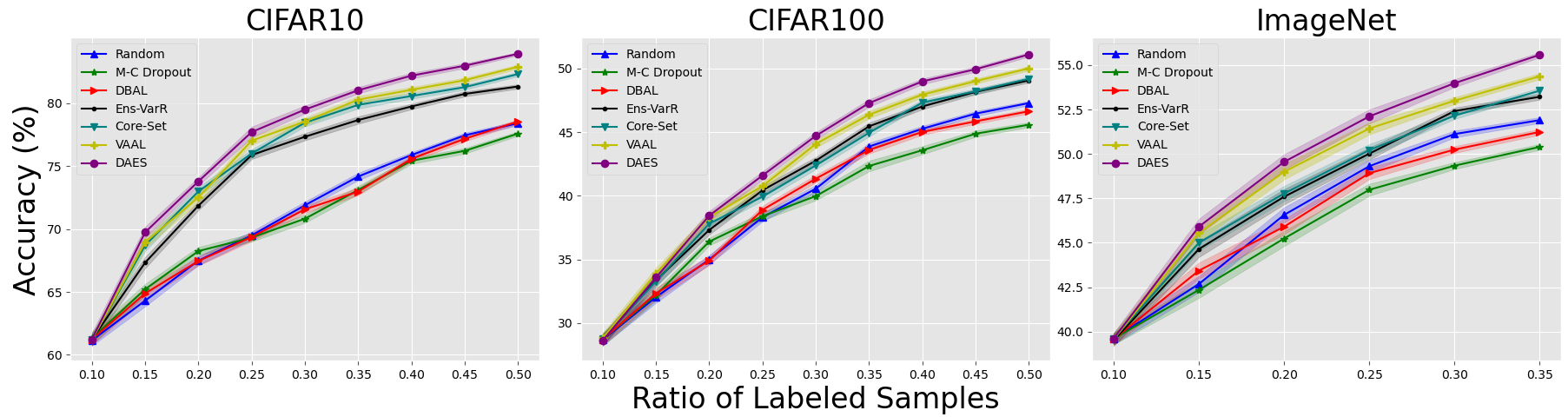}
		\caption{ Accuracy vs ratio of labeled samples from CIFAR10, CIFAR100 and ImageNet datasets.
		\label{Fig.0}}}
\end{figure*}
\subsection{DAES performance comparison}
In this section we explain the immediate results of experiments on MNIST, CIFAR10/100 and ImageNet in two comparing scenarios, namely, AL model trained from scratch, and AL on self-trained model.\\ 
\textbf{1. Trained from scratch:} The conventional protocol is training from scratch.

Our results on MNIST is on par with VAAL and Core-Set where all three approaches attained $99+\%$ accuracy with 1000 samples ($1.67\%)$ of the data. Ens-VarR, DBAL, M-C Dropout and Random baselines achieved $97.81\pm0.12$, $97.55\pm0.18$, $97.26\pm0.14$, and $95.2\pm0.23$

On CIFAR10 as shown in Fig.\ref{Fig.0}, our framework tends to outperform other baselines including VAAL upon using more than $15\%$ of the data, while the difference grows by adding more labeled samples. Our approach attains mean accuracy of $82.98\$$ and $83.93\%$ upon using $40\%$ and $50\%$ of the data respectively, whereas Top-1 accuracy using $100\%$ of data is $93.27\%$. Second and third highly performant methods using half of the data are VAAL and Core-Set with $82.89\%$ and $82.31\%$ respectively. While Ens-VarR remains fairly competitive, M-C dropout as well as DBAL are evidently underperforming.

On CIFAR100 also our method starts to outperform competitive VAAL and Core-Set approaches upon using $20+\%$ of the data. The accuracy difference swiftly grows by adding more samples to the point that upon using $50\%$ of data, our method outperforms VAAL and Core-Set by $51.33\%$ to $50.01\%$ and $49.03\%$. Note that the Top-1 accuracy using full data is $75.43\%$. As it is clear, due to larger number of classes, Core-Set experienced performance degradation down to performing on par with Ens-VarR . 

\begin{figure*}[t]
\centering{
\includegraphics[scale=0.27]{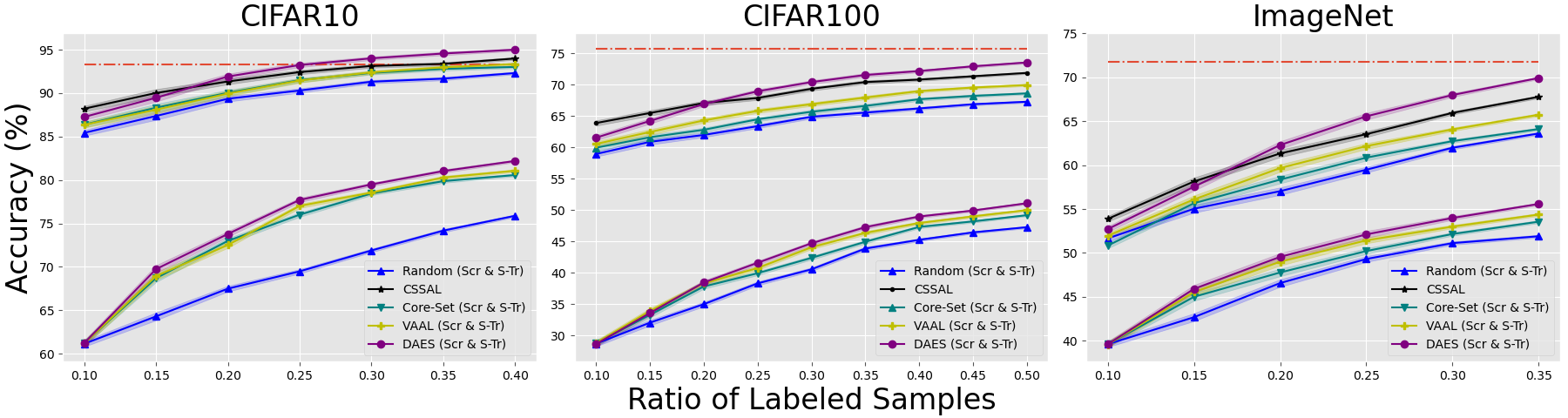}
		\caption{ Accuracy vs ratio of labeled samples from CIFAR100 and ImageNet datasets on SSL boosted networks. Top groups: results with proposed  training protocol using  SSL; Lower group: results with training without SSL. Red line denotes results using supervised learning with the full labeled data. 
		\label{Fig.1}}}
\end{figure*}

\textbf{Performance on dataset at scale:} On ImageNet as a large and more challenging dataset of $1.2+$ million samples of 1000 classes, our method patently outperforms former baselines upon using $15\%$ or more of data. Compared to Top-1 mean accuracy of $71.8\%$ using whole data, we achieve mean accuracy of $55.57\%$ upon using only $35\%$ of data, which is a $1.2\%$ improvement over VAAl, (while VAAL offers only less than $1\%$ improvement over its former baseline, Core-Set using $35\%$ of data). Our method improves over Random baseline by mean accuracy of $3.67\%$. Similar to their performance on CIFAR10 and CIFAR100, Bayesian techniques, i.e., DBAL and M-C dropout, slightly underperform Random baseline.\\
\textbf{2. Self-Training:} We also evaluated the proposed use of self-supervised knowledge distillation \cite{xu2020knowledge} from unlabeled data 
as a general technique to further improve the model training process for AL methods. 
There are two objectives here. 
First, to provide a fair comparison of this SSL+AL approach when applied on our proposed DAES, and three other AL baselines (namely, VAAL, Randon and Core-Set), against two approaches \cite{guo2021semi,gao2020consistency} that take advantage of knowledge distillation of unlabeled data. Second, to show that the SSL+AL protocol establishes a new standard training protocol for deep AL regardless of the underlying principle. As shown in Fig. \ref{Fig.1}, extensive experiments on Random baseline, VAAL, Core-Set and DAES on CIFAR10, CIFAR100 and ImageNet consistently confirm the performance jump due to SSL-wise leveraging of unlabeled data 
while still using a small percentage of labeled data.
Aside from bringing some accuracy jump to VAAL, Core-Set and Random baseline, this allows our framework to outperform CSSAL \cite{gao2020consistency} on CIFAR10, CIFAR100 and ImageNet by using $18+\%$, $20+\%$ and $17\%$ of data. 
As can be observed, the performance of our method on all three datasets rivals Top-1 mean accuracy attained by supervised learning (having the whole data labeled, denoted by the red line in the figure). On CIFAR10  and only using $40\%$ of data (labeled), all approaches except for Random acquisition perform above Top-1 mean accuracy of  $93.27\%$. On CIFAR100 ($50\% $ labeled) and ImageNet ($35\%$ labeled) all methods are competitive to supervised Top-1 mean accuracy, with our method (DAES) achieving a mean accuracy of $73.55\%$ ( compared to $75.81\%$) and $69.92\%$ (compared to $71.80\%$ ). Finally, compared with a recent baseline on semi-supervised learning, REVIVAL proposed in \cite{guo2021semi}, on CIFAR10 and using $40\%$ of the data, our framework performs on par with REVIVAL. On CIFAR100 our approach (using $35\%$ of the data) performs on par with REVIVAL (using $25\%$ of the data).
\section{Ablation study and investigative scenarios}
In this section 
we discuss our  ablation studies to assess the effect of model size on tuning the trade-off between performance and model capacity/complexity, DAES behaviour with deeper networks, 
and finally incremental training.
For all methods, we used Variation Ratio as acquisition function, as it is empirically proved to be the most effective query strategy in the literature \cite{gal2017deep,beluch2018power}. 
\subsection{DAES model size}
\begin{figure*}[t]
	\centering{
	\includegraphics[scale=.19]{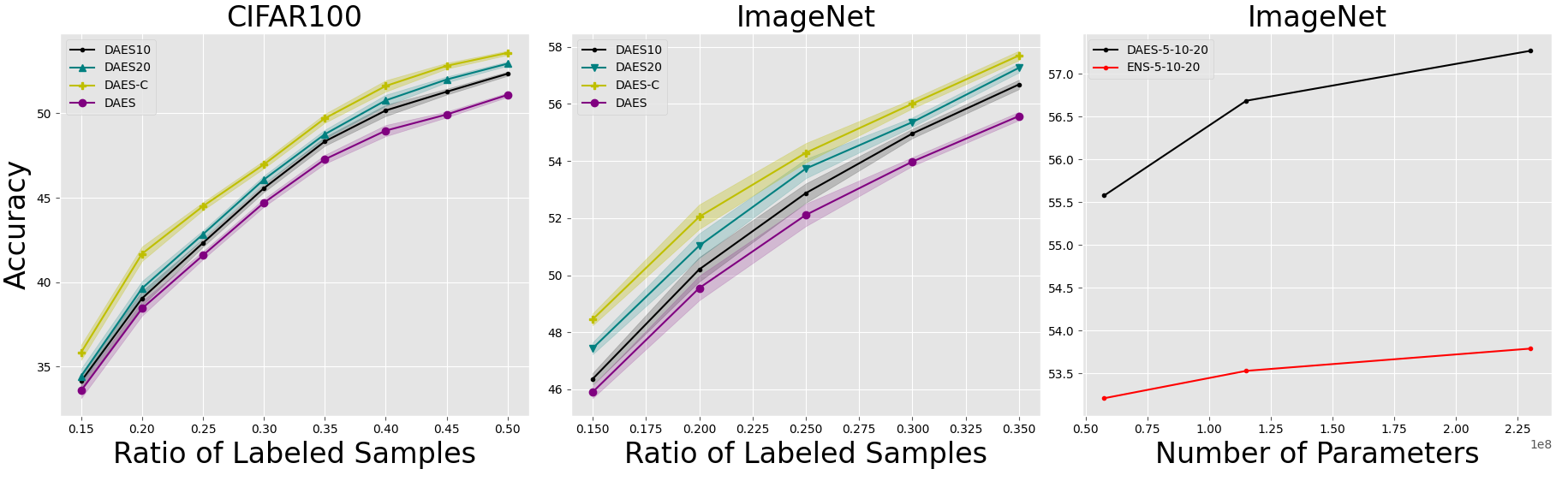}
		\caption{ Accuracy with different size of DAES and cumulative training (left and middle); and the slope of the curve representing a tunable trade-off for DAES (right). 
		\label{Fig.111}}}
\end{figure*}

One main advantage of DAES is that it can provide higher accuracy by enlarging the ensemble. Ens-VarR enjoys a performance boost only when changing the 1-member ensemble to ensemble with more than one member. Unlike Ens-VarR 
which lacks a malleable trade-off between  computational over-head and performance, meaning that adding reasonably more classifiers to the ensemble does not lead to a  proportional increase in performance,
we empirically assess how larger ensembles 
provide desirable improvement in accuracy for DAES. As shown in Fig. \ref{Fig.111}, DAES-10 with 5 additional classifiers (total of 10), approximately doubles the former accuracy improvement on ImageNet dataset (3 times the accuracy improvement that VAAL adds to Core-Set under the same experimental setting). This is while DAES-20 with 20 classifiers brings $180+\%$ improvement over 5-member DAES. Similar experiments on CIFAR100 also confirm the 
proportional improvement. CIFAR10 however enjoys relatively smaller accuracy enhancement compared with the other  datasets. We suspect that the underlying cause of the source of the improvement  could be due to two separate reasons. First, adding more classifiers positively impacts the model's capacity on efficient sample selection. Second, training the model on full training budget, allows classifiers to individually specialize in diverse feature representation and accordingly yields to a better generalization at test time, compared to a model with fewer classifiers. The former explanation could be intuitively conceived as the performance/behaviour spectrum of ensembles with $1, 2, ..., N$ member(s) over test time.
In all experiments, cumulative training using the union of chosen samples by DAES, VAAL and Core-Set performs slightly better than DAES-20 except for CIFAR10. We see this as indicative of the superior effect of training budget size over model capacity on overall performance in this setting. Also as shown in Fig \ref{Fig.111}, compared with DAES, 
training VAAL with the same cumulative training budget led to lower accuracy -- a  clear contrast of the models' capacities.   
\subsection{DAES with deeper networks}
We closely watch the training behavior of DAES with deeper networks such as ResNet50 and ResNet101. As an occasionally observed drawback, DAES built on very deep networks
such as ResNet101, tends to take much longer convergence time than expected, or could even fail to converge. As a remedy, we found it helpful to initially pre-train the networks (or blocks) separately using initial training budget, and then train the ultimate ensemble built using pre-trained blocks. Applying this simple trick ensures the convergence of DAES.

On ImageNet and using $35\%$ of data, DAES-5 with ResNet101 brings only approximately $1\%$ mean accuracy improvement over DAES-5 with ResNet18 which is less than the improvement provided by DAES-10 with ResNet18. The same behavior was observed with  CIFAR100. We suspect that adding more members (ensembles) to DAES leads to more improvement than replacing the blocks with deeper CNNs. 
\subsection{Incremental training and tunable accuracy/cost trade-off}
\textbf{A. Incremental training:} In a standard AL experimental setting, after updating the training set, the next training iteration starts from \textbf{scratch} (here for some 100 epochs). However, we investigate incremental training of models (VAAL, Core-Set and DAES) in which models are trained under much fewer number of epochs at each iteration while in next iteration rather than restarting the training, training continues. Specifically, we train the model for 20 epochs (formerly 100 epochs) with initial budget. Then after each data acquisition, the model first is trained on newly selected samples for as many epochs as former samples trained over, and next, the model will be trained on the updated training set for 20 epochs. This is to utilize a not fully trained model to leverage its current data representation for sample selection. Interestingly, we find that this could be a trick to speed up the active learner. Briefly, DAES, VAAL and Core-Set experience respective performance degradation of $(1.07\pm0.12)\%$, $(1.39\pm0.14)\%$ and $(1.51\pm0.11)\%$, respectively. In this setting Core-Set offers only $0.14\%$ mean accuracy gain over random acquisition under previous setting. Our analysis on time complexity briefly shows that the ratio (to DAES) of average consumed time for one iteration of sample selection for DAES, VAAL, Core-Set and DAES with incremental training were 1, 0.57, 3.78, and 0.24 respectively.\\
\textbf{B. Tunable trade-off:}
Consistent with the results in \cite{beluch2018power}, we could not see much accuracy improvement with increasing the ensemble size in classical ensemble-based methods as shown in Fig. \ref{Fig.111} (right figure). In other words, such classical methods do provide a tunable trade-off between accuracy and computational overhead. A satisfactory accuracy would be attained using 5 members, and increasing the number of members does not seem to proportionally improve the performance. 
However, active ensemble sampling showed a much robust performance in terms of exploiting more model capacity by adding more members to the ensemble. In fact, the Bayesian nature of active ensemble sampling in conjunction with its ensemble-designed structure allows achieving much higher accuracy by enlarging the ensemble at the cost of a proportional increase in computational overhead.
\section{Conclusion and future work}
In this paper, we introduced deep active ensemble sampling (DAES) inspired by an efficient approximation of Thompson sampling in order to combine the advantages of uncertainty-based and geometric-based approaches into one unified framework. We also examine a new training protocol formed on self-supervised knowledge distillation from unlabeled data on four baselines in order to confirm its effectiveness. Our framework is  assessed on four benchmark datasets in two experimental settings to establish a new baseline. Finally we pose a few scenarios aiming for analysing DAES. We leave further theoretical and empirical analyses on DAES with asymmetric architectures for future research.
\section{Acknowledgement}
This work is supported in part by grants from the US National Science
Foundation (Award $\#1920920$, $\#2125872$).
\bibliographystyle{splncs}
\bibliography{egbib}

\end{document}